\documentclass[lettersize,journal]{IEEEtran}
\usepackage{amsmath,amsfonts}
\usepackage{algorithmicx}
\usepackage{algorithm}
\usepackage{array}
\usepackage[caption=false,font=normalsize,labelfont=sf,textfont=sf]{subfig}
\usepackage{epsfig}
\usepackage{textcomp}
\usepackage{stfloats}
\usepackage{url}
\usepackage{verbatim}
\usepackage{graphicx}
\usepackage{cite}
\usepackage{svg}
\usepackage{soul, color, xcolor}
\usepackage{bibentry}

\usepackage{multirow}
\usepackage{wasysym}

\usepackage{algpseudocode}

\usepackage{etoolbox}

\begin{document}

\title{A Unified Framework for Iris Anti-Spoofing: Introducing Iris Anti-Spoofing Cross-Domain-Testing Protocol and Masked-MoE Method}

\author{
    Hang Zou,
    Chenxi Du,
    Ajian Liu,
    Yuan Zhang,
    Jing Liu,
    Mingchuan Yang,
    Jun Wan,~\IEEEmembership{Senior Member, IEEE},
    Hui Zhang,
    Zhenan Sun~\IEEEmembership{Senior Member, IEEE}.

\thanks{

Hang Zou and Chenxi Du contributed equally to this work. The corresponding author is Hui Zhang.

Hang Zou, Yuan Zhang, and Mingchuan Yang are with the China Telecom Research Institute, Beijing 102209/ Shanghai 200122, China (e-mail: zouh3@chinatelecom.cn; zhangy666@chinatelecom.cn; yangmch@chinatelecom.cn).

Chenxi Du is with the Southern University of Science and Technology (SUSTech), Shenzhen 518055, China, the Shenzhen Institutes of Advanced Technology (SIAT), Shenzhen 518055, China, the Shenzhen University of Advanced Technology (SUAT), Shenzhen 518107, China (e-mail: cxdu025@163.com).

Ajian Liu and Jun Wan are with the State Key Laboratory of Multimodal Artificial Intelligence Systems (MAIS), Institute of Automation Chinese Academy of Sciences
(CASIA), Beijing 100190, China, and also with the School of Computer Science and Engineering, Macau University of Science and Technology, Macau, China (e-mail: ajianliu92@gmail.com; jun.wan@ia.ac.cn).

Jing Liu and Hui Zhang are with the Tianjin University of Science \& Technology, Tianjin 300457, China (e-mail: jing.liu@tust.edu.cn; zhanghui2022@tust.edu.cn).

Zhenan Sun is with the Center for Research on Intelligent Perception and Computing, State Key Laboratory of Multimodal
Artificial Intelligence Systems, Institute of Automation, Chinese Academy of Sciences, Beijing 100190, China (e-mail: znsun@nlpr.ia.ac.cn).

}

}

\markboth{Journal of \LaTeX\ Class Files,~Vol.~14, No.~8, August~2021}%
{Shell \MakeLowercase{\textit{et al.}}: A Sample Article Using IEEEtran.cls for IEEE Journals}


\maketitle

\begin{abstract}
Iris recognition is widely used in high-security scenarios due to its stability and distinctiveness.
However, iris images captured by different devices exhibit certain and device-related consistent differences, which has a greater impact on the classification algorithm for anti-spoofing.
The iris of various races would also affect the classification, causing the risk of identity theft.
So it is necessary to improve the cross-domain capabilities of the iris anti-spoofing (IAS) methods to enable it more robust in facing different races and devices.
However, there is no existing protocol that is comprehensively available.
To address this gap, we propose an Iris Anti-Spoofing Cross-Domain-Testing (IAS-CDT) Protocol, which involves 10 datasets, belonging to 7 databases, published by 4 institutions, and collected with 6 different devices.
It contains three sub-protocols hierarchically, aimed at evaluating average performance, cross-racial generalization, and cross-device generalization of IAS models.
Moreover, to address the cross-device generalization challenge brought by the IAS-CDT Protocol, we employ multiple model parameter sets to learn from the multiple sub-datasets.
Specifically, we utilize the Mixture of Experts (MoE) to fit complex data distributions using multiple sub-neural networks.
To further enhance the generalization capabilities, we propose a novel method Masked-MoE (MMoE), which randomly masks a portion of tokens for some experts and requires their outputs to be similar to the unmasked experts, which can effectively mitigate the overfitting issue of MoE.
For the evaluation, we selected ResNet50, VIT-B/16, CLIP, and FLIP as representative models and benchmarked them under the proposed IAS-CDT Protocol.
The results show our proposed method MMoE with CLIP achieves the best performance, which demonstrates its generalization performance and its training effectiveness on our proposed IAS-CDT Protocol.
\end{abstract}


\section{Introduction}

\begin{figure}[t]
\centering
\includegraphics[width=\columnwidth]{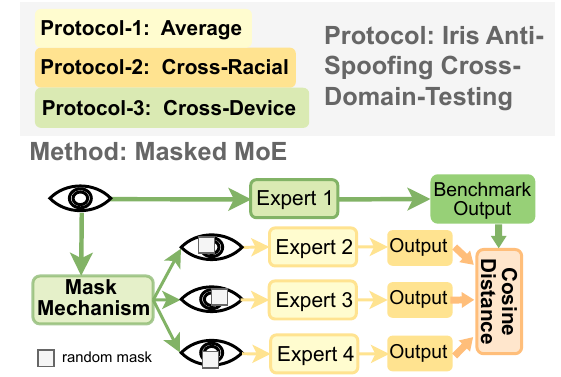} 
\caption{This figure shows the unified framework for iris anti-spoofing. Including the Iris Anti-Spoofing Cross-Domain-Testing (IAS-CDT) Protocol and Masked-MoE (MMoE) Method.}
\label{fig: main}
\end{figure}

\IEEEPARstart{T}{he} iris has large differences between each human, which is stable and distinguishable~\cite{daugman2009iris, agarwal2021presentation}, making it widely used in high-security scenarios.
Thus, it is necessary to ensure the security of the iris recognition system because it always relates to the property and critical areas~\cite{nigam2015ocular, perala2017princeton,gent2023cryptocurrency}.
Iris recognition systems are similar to other kinds of biometric detection systems that are vulnerable to spoofing, such as paper prints, screens, contact lenses, and plastic eyes~\cite{yadav2019synthesizing, yadav2020relativistic}.
Among those, contact lenses are complicated to detect because they are worn on real eyeballs and can fool the liveness detection algorithems~\cite{bowyer2014cosmetic, galbally2012iris}, which harm the security of the iris recognition systems~\cite{he2009new, daugman2004iris, von2005countermeasures, wei2008counterfeit}.
There are various types of contact lenses, most of which are transparent with some textures in public datasets.
It is intertwined with the original textures of the iris, making the identity difficult to classify~\cite{marcel2023handbook}.

\begin{table*}[ht]
\fontsize{10pt}{10pt}\selectfont
\setlength{\tabcolsep}{3pt}
\renewcommand{\arraystretch}{1.2}
\centering
\begin{tabular}{|c|c|c|c|c|c|c|}

\hline
DataBase                                                                                         & Dataset            & Devices      & Nickname & \begin{tabular}[c]{@{}c@{}}Num of Train \\ (Real / Fake)\end{tabular} & \begin{tabular}[c]{@{}c@{}}Num of Test\\ (Real / Fake)\end{tabular} & Num of Total \\ \hline
CASIAH100~\cite{sun2013iris}   & CASIAH100  & H100 & H & 4806 / 592 & 1198 / 148 & 6744         \\ \hline

\begin{tabular}[c]{@{}c@{}}Clarkson 2013 for \\ LivDet-Iris 2013\end{tabular}~\cite{yambay2017livdet}
& Clarkson2013Dalsa  & DALSA    & Da & 270 / 400 & 246 / 440 & 1356         \\ \hline

\multirow{2}{*}{\begin{tabular}[c]{@{}c@{}}Clarkson 2015 for \\ LivDet-Iris 2015\end{tabular}~\cite{yambay2017livdet}}
& Clarkson2015Dalsa  & DALSA    & Db & 700 / 873 & 378 / 558 & 2509         \\ \cline{2-7}
& Clarkson2015LG2200 & LG2200   & La & 450 / 540 & 378 / 576 & 1944         \\ \hline

\begin{tabular}[c]{@{}c@{}}Clarkson 2017 dataset \\ for LivDet-Iris 2017\end{tabular}~\cite{yambay2017livdet}
& Clarkson2017LG2200 & LG2200   & Lb & 2469 / 1122 & 1485 / 765 & 5841      \\ \hline

IFVE~\cite{zhang2021local}    & IFVEAI1000 & AI1000 & F  & 20000 / 20000 & 5000 / 5000 & 50000        \\ \hline

\multirow{2}{*}{\begin{tabular}[c]{@{}c@{}}Notre Dame 2013 for \\ LivDet-Iris 2013\end{tabular}~\cite{doyle2014notre}}
& ND2013LG4000       & LG4000 & Ga  & 2000 / 1000 & 800 / 400 & 4200         \\ \cline{2-7}
& ND2013AD100  & AD100 sensor & Aa   & 400 / 200  & 200 / 100 & 900          \\ \hline

\multirow{2}{*}{\begin{tabular}[c]{@{}c@{}}Notre Dame 2017 for \\ LivDet-Iris 2017\end{tabular}~\cite{yambay2017livdet}}
& ND2017LG4000  & LG4000      & Gb   & 600 / 600  & 900 / 900 & 3000         \\ \cline{2-7}
& ND2017-CDAD100 & AD100 sensor & Ab  & 0   & 900 / 900  & 1800         \\ \hline

\end{tabular}
\vspace{1.0em}
\caption{This Table shows the scale of the sub-datasets employed in the Iris Anti-Spoofing Cross-Domain-Testing (IAS-CDT) Protocol, including the numbers of real and fake samples in both training and testing sets. The first letters of the nicknames denote the collection devices, while the second letters denote the order to distinguish.}
\label{Tab: datasets}
\end{table*}

Extensive previous works combat contact lens attacks, primarily categorized into hardware-based, software-based, and challenge-response approaches~\cite{marcel2023handbook}.
Both hardware-based and challenge-response approaches need extra devices when detecting.
The hardware-based one needs special sensors, in addition to standard iris sensors, to measure the special characteristics of eyes, in contrast, the challenge-response one utilizes extra devices to stimulate the eyes and detect the spoofing samples by comparing the captured eyes' responses.
Unlike these two, software-based approaches merely use static features extracted from samples acquired by standard iris sensors.
For traditional software-based methods, the study mainly focused on features, they designed different extractors to get the distinctive features from iris images and used them for classification.
There are several represent works such as the sharpness of edges~\cite{wei2008counterfeit}, gray level value~\cite{he2007statistical}, Local Binary Pattern (LBP)~\cite{zhang2010contact, gupta2014iris, he2009efficient}, Binary Statistical Image Features (BSIF)~\cite{raghavendra2014presentation}, Scale-Invariant Feature Transform (SIFT)~\cite{zhang2011learning, sun2013iris}, a fusion of 2D and 3D features~\cite{silva2015approach}, Local Phase Quantization (LPQ)~\cite{gragnaniello2015investigation}, etc.
Deep-learning-based methods mainly amend the structure of the models, compared to previous methods, instead of modifying the extraction and design of features, which simplifies the feature extraction process and allows models to learn the critical differences between real samples and the spoofing ones by updating the parameters themselves, the advantage of deep-learning-based methods.
For example, the Attention Learning~\cite{fang2021iris, chen2021explainable} methods aim to fuse multi-scale features, which could improve the performance of the CNN framework, make the models concentrate more on the global features, and alleviate the overfitting caused by CNN focusing on local features.
The Adversarial Learning~\cite{yadav2019synthesizing, yadav2021cit, zou2018generation} aims to generate unseen iris samples for anti-spoofing models to improve their generalization, and the convolutional structure~\cite{safaa2020deep} aims to extract the critical features using neural networks.

Nowadays, the development of Large Language Models (LLMs) provides a novel way to deal with these Biometric Anti-Spoofing problems, which could use the pre-trained models to extract the features of the samples.
Some novel works show good performance on the Face Anti-Spoofing (FAS) task with LLMs, for example, the FLIP~\cite{srivatsan2023flip} proposed by Srivatsan et al. combines the visual transformers and language guidance based on the CLIP~\cite{clip}, and fine-tunes it with different strategies to form three variants: FLIP-Vision (FLIP-V), FLIP-Image-Text Similarity (FLIP-IT), and FLIP-Multimodal-Contrastive-Learning (FLIP-MCL).
The Class-Free Prompt Learning (CFPL)~\cite{liu2024cfpl} proposed by Liu et al. introduces two lightweight transformers, Content Q-Former (CQF) and Style Q-Former (SQF), to learn different semantic prompts based on content and style features. It leverages the large-scale vision-language model CLIP to dynamically adjust the classifier’s weights, significantly enhancing the generalization of extracted features.
Zhang et al. introduced a multimodal fine-grained CLIP (MFCLIP)~\cite{zhang2024mfclip}, which designed an innovative Sample Pair Attention (SPA) module, which can adaptively highlight relevant negative sample pairs, optimizing cross-modal feature alignment, greatly improving the generalization ability for detecting diffusion-generated facial fakes.

Though the deep-learning-based methods have good results in experiments, they are trained on specific datasets, which makes it hard to cope with the situation once the collection devices are changed.
The previous study proves that the image-imaging differences between each iris collection device are obvious and fixed.
The near-infrared fill light bands, camera filters, collection distances, and ambient light, make a lot of impact on iris acquisition which requires near-infrared presentation~\cite{xiao2013coupled}, which could bring negative impacts on iris anti-spoofing (IAS) models in practice.
This means that the cross-domain ability of the IAS methods is necessary, thus the security could be maintained more effectively.
However, there are no suitable datasets or protocols to train and evaluate IAS models' generalization on cross-domain adaptation~\cite{yadav2024synthesizing}.

To address this gap, we propose an Iris Anti-Spoofing Cross-Domain-Testing (IAS-CDT) Protocol, which involves 10 datasets, belonging to 7 databases, published by 4 institutions, and collected with 6 different devices.
It contains three sub-protocols hierarchically.
Protocol-1 evaluates the model’s average performance.
Protocol-2 evaluates the model’s generalization performance of cross-racials.
Protocol-3 evaluates the model’s generalization performance of cross-devices.
Section \ref{Section: protocol} provides more details of this IAS-CDT protocol.

The proposal of the IAS-CDT Protocol brings the challenge of integrating multiple sub-datasets with different races and devices.
To tackle it, we employ multiple parameter sets to learn from the various subsets by using the Mixture of Experts (MoEs)~\cite{jacobs1991adaptive} to fit complex data distributions with multiple sub-neural networks.
The concept of MoEs has evolved over at least 30 years~\cite{fedus2022review}, originating with the work of Jacobs et al. in 1991~\cite{jacobs1991adaptive}.
Initially, MoEs were introduced as a supervised learning framework known as "adaptive mixtures of local experts," where multiple independent networks collaboratively processed subsets of training samples.
More recently, the application of MoEs has gained prominence through the work of Shazeer et al., who introduced the Sparsely-Gated Mixture-of-Experts Layer~\cite{shazeer2017outrageously}.
This layer enabled sparse gating and token-level processing, paving the way for scalability in large language models (LLMs).
Based on this, Lepikhin, Dmitry, et al. proposed GShard~\cite{lepikhin2020gshard} represents a significant milestone as the first to implement MoE within a Transformer framework, enhancing efficiency in large-scale training, improving LLMs model performance on sparse tasks.
Since then, more and more MoE-related works have been proposed, such as V-MoE~\cite{riquelme2021scaling}, ST-MoE~\cite{zoph2022st}, Switch Transformer~\cite{fedus2022switch}, LIMoE~\cite{mustafa2022multimodal}, DeepseekMoE~\cite{dai2024deepseekmoe}, MH-MoE~\cite{huang2024mh} and etc..

However, some previous works have found that sparse MoE would bring overfitting problems into models~\cite{shazeer2017outrageously, chen2023sparse, elbayad2022fixing, lepikhin2020gshard}.
It may be caused by the low utilization of parameters in sparse MoEs, which means that single or several experts integrate multi-tasks and other experts learn redundant knowledge because only parts of the experts deal with the input, and the parameters of the unused experts are not updated.
At the same time, overfitting is also the main challenge of the iris anti-spoofing task, which is caused by its high structural homogeneity, leading to the differences between live iris and fake iris mainly in their fine textures.
Thus, alleviating the overfitting and improving the generalization of the model is necessary.
In that case, we introduce a novel Masked-MoE (MMoE) method to enhance the generalization capabilities and reduce the passive overfitting impacts brought by MoE.
It has two contributions: Mask Mechanism to alleviate overfitting, and Cosine Distance Loss to further improve the model's generalization.
Specifically, the Mask Mechanism masks parts of the input tokens for several experts and allows one expert to see all tokens with complete information.
Then the Cosine Distance Loss requests the output of masked experts to be similar to the unmasked experts as shown in Figure~\ref{fig: main}.
It is inspired by MAE~\cite{he2022masked}, and we refer to this masking method to improve the generalization to deal with the inherent overfitting problem of MoE.
The Mask Mechanism could reduce the overfitting problem by dropping out a reasonable rate of the input iris features, while the Cosine Distance improves the model's generalization by driving different Experts in MoE to adapt unknown knowledge by closing the cosine distance between masked features and unmasked features.
Section Proposed Method provides the details of our MMoE method.

As we mentioned before, the development of LLMs provides a novel way to deal with Biometric Anti-Spoofing problems.
Since CLIP~\cite{clip} has already performed its powerful and competitive ability in the FAS field, for example UniAttackDetection~\cite{fang2024unified}, La-SoftMoE~\cite{zou2024softmoe}, CPL-CLIP~\cite{zhang2024cpl}, we choose it as the backbone of our proposed Iris Anti-Spoofing model.
CLIP~\cite{clip} is the visual-linguistic dual-modality pre-trained model.
It has two encoders, an image encoder and a text encoder, the contrastive learning loss is employed to drive the training of the two encoders mapping the feature of different modalities to the same latent space with similar information to close the distance of image embedding and text embedding.

We selected ResNet50~\cite{he2016deep}, ViT-B/16~\cite{ViT},  CLIP~\cite{clip}, and FLIP~\cite{srivatsan2023flip} as representative models to evaluate our contributions and benchmarked them on the IrisGeneral dataset.
Experimental results show that our proposed MMoE with CLIP has strong generalization abilities and achieves the best performance on both three protocols, section~\ref{Section: Experiment} provides more details.

To sum up, the main contributions in this paper are shown as follows:

\begin{itemize}
    \item
    \textbf{We proposed a unified framework for Iris Anti-Spoofing,} including an Iris Anti-Spoofing Cross-Domain-Testing (IAS-CDT) Protocol and a Masked-MoE Method. To address the gap of no suitable datasets or protocols to train and evaluate
IAS models’ generalization on cross-domain adaptation. And improve the competition of iris recognition systems in the anti-spoofing area.

    \item
    \textbf{Protocol: Iris Anti-Spoofing Cross-Domain-Testing.} The IAS-CDT Protocol is proposed with three sub-protocols that could respectively evaluate the average, cross-racial, and cross-device performance of the iris-antispoofing task.
    \item
    \textbf{Method: Masked-MoE.}The MMoE method is proposed with the Mask Mechanism and the Cosine Distance Loss, which efficiently alleviates the problem of MoE method overfitting and improves the models' generalization.
\end{itemize}

\section{Related Work}

This section introduces several previous works on Iris Anti-Spoofing (IAS), Cross-domain Adaptation, Cross-domain Adaptation in Biometric Anti-Spoofing, and Mixture-of-Experts (MoEs). 

\subsection{Iris Anti-Spoofing (IAS)}
\begin{figure}[h]
\centering
\includegraphics[width=0.8\columnwidth]{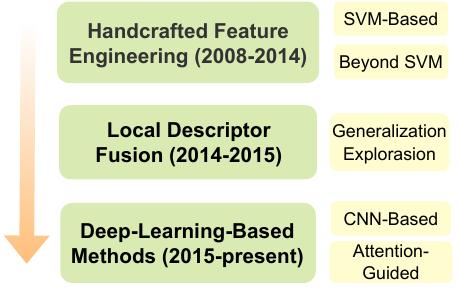}
\caption{This figure shows the development of the iris anti-spoofing (IAS) methods, mainly separated into three phases.}
\label{fig: develop}
\end{figure}

The evolution of iris anti-spoofing (IAS) methods has experienced three distinct phases, each marked by different feature engineering paradigms. As shown in Fig~\ref{fig: develop}, the phases are Handcrafted Feature Engineering, Local Descriptor Fusion, and Deep-Learning-Based methods respectively.

\textbf{The first phase, Handcrafted Feature Engineering (2008-2014)}, saw some IAS research focus on designing manual texture descriptors for SVM-based classification, while other research explored new classifiers to replace SVM due to its costly computation.
Within SVM-based research, there are still many different types of approaches that stood out. There are some GLCM-based methods.
For instance, Wei et al.~\cite{wei2008counterfeit} proposed a texture analysis approach for fake iris detection, incorporating features like iris edge sharpness, Iris-Texton features, and GLCM-based features.
Alonso et al.~\cite{alonso2014fake} presented an iris anti-spoofing method using GLCM for feature extraction paired with an SVM classifier, which saved time and reduced complexity.
Methods based on the Wavelet Transform also performed well on IAS field.
He et al.~\cite{he2009new} introduced an anti-spoofing method based on Wavelet Packet Transform (WPT) and SVM, effectively capturing distinctive features between fake and genuine irises.
Zhang et al.~\cite{zhang2010contact} developed an iris anti-spoofing method using SIFT descriptors and weighted LBP (w-LBP), with SVM as the classifier.
Another dominant approach was the Hierarchical-Visual-Codebook-based method.
Sun et al.~\cite{sun2013iris} proposed an iris image classification method based on the Hierarchical Visual Codebook (HVC),which included iris image preprocessing, low-level visual feature extraction, statistical iris image representation based on the HVC model, and iris image classification.

Some research in this phase also focused on classifier innovation beyond SVM.
He et al.~\cite{he2009efficient} proposed an anti-spoofing decision model based on LBP and Adaboost learning, which effectively differentiated genuine and fake iris images, showing strong performance against specific spoofing methods like contact lenses.

\textbf{The second phase, Local Descriptor Fusion (2014-2015)}, saw researchers refine feature granularity for spoofing iris detection and explore models' generalization ability on cross-device tasks.
Gupta et al.~\cite{gupta2014iris} introduced an iris anti-spoofing approach using SVM as the classifier with traditional feature extraction methods like LBP, Histogram of Oriented Gradients (HOG), and GIST.
Raghavendra et al.~\cite{raghavendra2014presentation} developed a Presentation Attack Detection (PAD) method for face and iris anti-spoofing, employing linear SVM for decision-making.

\textbf{The third phase, Deep-Learning-Based methods (2015-present)}, divides IAS research into hybrid architectures and fully deep-learning-based methods. Hybrid architectures blended deep learning with manual features.
For example, Chen et al.~\cite{chen2021explainable} combined the advantages of the attention mechanism and CNN, proposing a novel method to detect the detection of iris presentation attacks, named Attention-Guided PAD (AG-PAD).
Fang et al.~\cite{fang2021iris} proposed an IAS method, called Attention-based Pixel-level Binary Supervision (A-PBS), based on attention mechanism and CNN as well.
Wu et al.~\cite{wu2024unsupervised} proposed an unsupervised domain adaptation transfer learning model to enhance the precision and generalization of iris anti-spoofing, leveraging contrastive learning to enhance characteristic characteristics of features.

While these methods have performed well in IAS, there's a lack of large-scale or wide-ranging generalization evaluation. The development of multi-modal pre-trained models and LLMs offers new ways to improve IAS model performance and generalization. Thus, we propose an IAS model based on CLIP and MoE (a key LLM component), evaluating its performance and generalization on our proposed IAS-CDT Protocol, which includes many datasets suitable for generalization evaluation.

\subsection{Cross-domain Adaption}

The evolution of domain adaptation methods in pattern recognition has been marked by two major paradigm shifts: Adversarial Learning Dominance and Feature Space Unification.

The Adversarial Learning Dominance paradigm was initiated by adversarial training methods, which align domain distributions through minimax optimization.
For example, Tzeng et al.~\cite{tzeng2017adversarial} proposed the Adversarial Discriminative Domain Adaptation (ADDA), an unsupervised domain generalization method that combines adversarial learning with discriminative feature learning.
A domain discriminator is trained to minimize the distance between the source and target domains, enabling direct application of the source classification model to target representations.
This combination of discriminative modeling and adversarial loss addresses limitations of traditional methods in handling large-scale domain shifts.
Long et al.~\cite{JAN} introduced Joint Adaptation Networks (JAN) to address data distribution disparities between source and target domains.
Unlike methods focusing only on marginal distribution alignment, JAN emphasizes joint distribution adaptation across multiple layers.
It has demonstrated excellent classification accuracy on various benchmark datasets, validating its effectiveness in deep transfer learning.

Feature Space Unification approaches, unlike adversarial learning methods that enhance model domain generalization, with the rise of large models bypass adversarial training by constructing domain-invariant feature spaces.
For example, Long et al.~\cite{DAN}  introduced a Deep Adaptation Network (DAN) to improve the transferability of deep neural networks in domain adaptation.
Ren et al.~\cite{ren2020learning} proposed Target Domain-Specific Classifier Learning (TSCDA) to address knowledge transfer challenges when source and target domain label spaces are inconsistent.
TSCDA includes two interdependent modules: Selective Weighted MMD (SWMMD) for partial feature alignment and PEers Assisted Learning (PEAL).
It minimizes the prediction discrepancy between the target and auxiliary classifiers, promoting collaborative learning among multiple target-specific classifiers and enhancing their discriminative power, effectively mitigating negative transfer and classifier bias in partial domain adaptation.

Although Adversarial Learning Dominance methods are theoretically flexible and universal, their training challenges make feature space unification methods more practical for real-world applications.

\subsection{Cross-domain Adaption in Biometric Anti-Spoofing}
Domain generalization presents a significant challenge for biometric anti-spoofing systems, as performance often degrades due to cross-domain discrepancies, including sensor variations, shifts in attack types, modality differences, and other factors.

Early approaches utilized deep learning architectures, particularly CNNs, to establish baseline generalization capabilities.
Silva et al.~\cite{silva2015approach} employing a CNN to construct deep image representations and a fully connected layer with softmax regression for classification, and demonstrated substantial performance improvements in cross-domain tests.
Safaa et al.~\cite{safaa2020deep} focused on PAD for face and iris modalities, featuring a deep CNN with multi-layer convolutional and pooling architecture, and produced binary classification results through fully connected layers.
Boyd et al.~\cite{boyd2021cyborg} presented CYBORG, which integrates human-annotated saliency maps into the loss function.
It achieved higher accuracy on unseen samples compared to traditional methods.

To overcome CNN limitations in capturing global dependencies, attention mechanisms emerged as a key advancement.
Fang et al.~\cite{fang2021iris} developed the Attention-based Pixel-level Binary Supervision (A-PBS) network, addressing CNN-based PAD method deficiencies like limited local feature capture and overfitting.
The A-PBS model outperformed DenseNet and PBS in most cross-database scenarios, validating its generalization capability.
Chen et al.~\cite{chen2021explainable} proposed an explainable Attention-Guided PAD (AG-PAD) for iris presentation attack detection, integrating attention mechanisms to enhance CNN performance and provide visual interpretability.

The rise of vision-language models spurred a paradigm shift toward multimodal domain adaptation.
Zhang et al.~\cite{cplclip} introduced CPL-CLIP, that uses dual-branch prompt learning to disentangle cross-modal domain shifts, which enables deployment-robust generalization without target domain knowledge.
Fang et al.~\cite{fang2024vl} proposed a visual-language framework (VL-FAS) for cross-domain face anti-spoofing, achieving domain-invariant features through language-driven semantic alignment.
Hu et al.~\cite{hufine} presented FGPL, a vision-language framework for cross-domain face anti-spoofing, enabling robust spoofing detection across unseen domains through joint optimization of text-visual cues and domain-invariant features.

Recent innovations emphasize domain-agnostic prompt engineering and frequency-space alignment.
Liu et al.~\cite{liu2024cfpl} proposed Class-Agnostic Prompt Learning (CFPL), advancing cross-domain face anti-spoofing by leveraging vision-language models to disentangle content and style semantics without domain labels.
Liu et al. also introduced FM-CLIP~\cite{liu_fmclip}, which enhances cross-domain adaptation by jointly aligning multimodal features in frequency and text-embedding spaces.
Guo et al.~\cite{guo2024style} developed Style-Conditional Prompt Token Learning (S-CPTL) for cross-domain face anti-spoofing, dynamically generating domain-invariant text prompts conditioned on visual style features to disentangle liveness semantics from domain shifts, advancing generalized biometric defense against domain drift.
\begin{figure*}[t]
\centering
\includegraphics[width=0.95\textwidth]{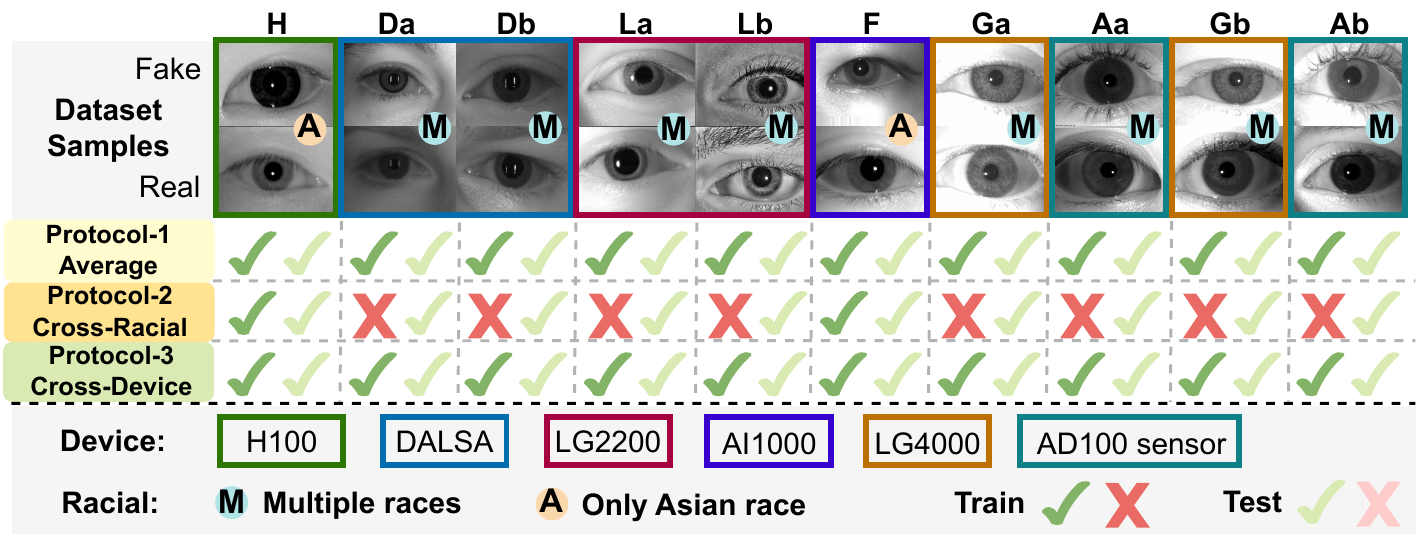} 
\caption{
This figure shows the Train / Test usage of each dataset of the IAS-CDT Protocol. In Protocol-1, all datasets' training and testing sets are involved. In Protocol-2, training sets are H and F, and testing sets involve all 10 subsets. In Protocol-3, for each sub-protocol, we choose one device to test and others to train.
And some image samples are shown, including 10 subsets with Real and Fake iris images.}
\label{fig: samples}
\end{figure*}

\subsection{Mixture-of-Experts (MoEs)}

The conceptual origins of MoE trace back to the work of Jacobs et al.~\cite{jacobs1991adaptive}, which proposed a modular architecture with competing expert networks for complex tasks like vowel recognition.
This design decomposes tasks into subtasks, reducing interference. The system includes multiple feedforward networks (experts) and a gating network.
Each expert receives the same input and produces outputs of the same size, while the gating network determines which expert to activate for each case.
Compared to traditional single networks, this modular design and expert selection mechanism allow for better handling of complex tasks and improved generalization.
This work is widely regarded as the seminal study of MoE.

Nowadays, MoE has regained attention due to its ability to handle complex tasks.
The sparsity-driven conditional computation started from the work of Shazeer et al.~\cite{shazeer2017outrageously}.
They introduced a novel sparse gating mechanism for MoE layers, aiming to significantly increase the model capacity of neural networks through conditional computation, addressing the limitations of traditional models in handling large-scale data. 
Since the Transformer-based models have performed well on various tasks,
Lepikhin et al.~\cite{lepikhin2020gshard} were the first to apply the MoE structure to the Transformer architecture.
This approach leverages a sparse activation MoE model with conditional computation, improving model quality without significantly increasing computational cost.

To further improve the stabilization of MoE, Zoph et al.~\cite{zoph2022st} proposed the ST-MoE model, which provides a systematic framework for sparse expert model design through architectural principles and routing algorithms.
The ST-MoE model employs a sparse expert architecture where each input is processed by only a small subset of experts, allowing for a significant increase in parameter count without a corresponding rise in computation.
Through the introduction of sparse expert architecture and a suite of stability techniques, the ST-MoE model demonstrates the potential of sparse models in transfer learning tasks.

To address the challenges brought by the sparse structure of MoE, fully differential MoE appears. Puigcerver et al.~\cite{puigcerver2023sparse} proposed a novel model, Soft MoE (Soft Mixture of Experts). Soft MoE assigns input tokens to expert slots using a weighted average computation, with each slot processed by a corresponding expert. By implementing a fully differentiable soft allocation mechanism, Soft MoE addresses common challenges in sparse MoE models, such as training instability, token dropout, poor scalability with expert numbers, and suboptimal fine-tuning performance, while maintaining the efficiency and capacity benefits of MoE.



\section{Iris Anti-Spoofing Cross-Domain-Testing Protocol}
\label{Section: protocol}

\subsection{Subsets Detail}
There are 10 subsets, belonging to 7 databases, published by 4 institutions, collected with 6 types of devices in the Cross-Testing Protocol.
Table~\ref{Tab: datasets} lists the information for each subset.
Figure~\ref{fig: samples} provides some real and fake iris image samples of each subset.

\noindent
\textbf{CASIAH100}~\cite{sun2013iris} with 6744 samples in total, collected by H100, belonging to database CASIAH100, contains 5398 samples in its training set and 1346 samples in its testing set. We call this dataset H in short.

\noindent
\textbf{Clarkson2013Dalsa}~\cite{yambay2017livdet} with 1356 samples in total, collected by DALSA, belonging to database Clarkson 2013 for LivDet-Iris 2013, contains 670 samples in its training set and 686 samples in its testing set. We call this dataset Da in short.

\noindent
\textbf{Clarkson2015Dalsa}~\cite{yambay2017livdet} with 2509 samples in total, collected by DALSA, belonging to database Clarkson 2015 for LivDet-Iris 2015, contains 1573 samples in its training set and 936 samples in its testing set. We call this dataset Db in short.

\noindent
\textbf{Clarkson2015LG2200}~\cite{yambay2017livdet} with 1944 samples in total, collected by LG2200, belonging to database Clarkson 2015 for LivDet-Iris 2015, contains 990 samples in its training set and 954 samples in its testing set. We call this dataset La in short.

\noindent
\textbf{Clarkson2017LG2200}~\cite{yambay2017livdet} with 5841 samples in total, collected by LG2200, belonging to the database Clarkson 2017 dataset for LivDet-Iris 2017, contains 3591 samples in its training set and 2250 samples in its testing set. We call this dataset Lb in short.

\noindent
\textbf{IFVEAI1000}~\cite{zhang2021local} with 50000 samples in total, collected by AI1000, belonging to database IFVE, contains 40000 samples in its training set and 10000 samples in its testing set. And we call this dataset F in short.

\noindent
\textbf{ND2013LG4000}~\cite{doyle2014notre}with 4200 samples in total, collected by LG4000, belonging to database Notre Dame 2013 for LivDet-Iris 2013, contains 3000 samples in its training set and 1200 samples in its testing set. We call this dataset Ga in short.

\noindent
\textbf{ND2013AD100}~\cite{doyle2014notre}with 900 samples in total, collected by AD100sensor, belonging to database Notre Dame 2013 for LivDet-Iris 2013, contains 600 samples in its training set and 300 samples in its testing set. We call this dataset Aa in short.

\noindent
\textbf{ND2017LG4000}~\cite{yambay2017livdet} with 3000 samples in total, collected by LG4000, belonging to database Notre Dame 2017 for LivDet-Iris 2017, contains 1200 samples in its training set and 1800 samples in its testing set. We call this dataset Gb in short.

\noindent
\textbf{ND2017-CDAD100}~\cite{yambay2017livdet} with 1800 samples in total, collected by AD100sensor, belonging to database Notre Dame 2017 for LivDet-Iris 2017, contains 0 samples in its training set and 1800 samples in its testing set. We call this dataset Ab in short.

\begin{figure*}[t]
\centering
\includegraphics[width=\textwidth]{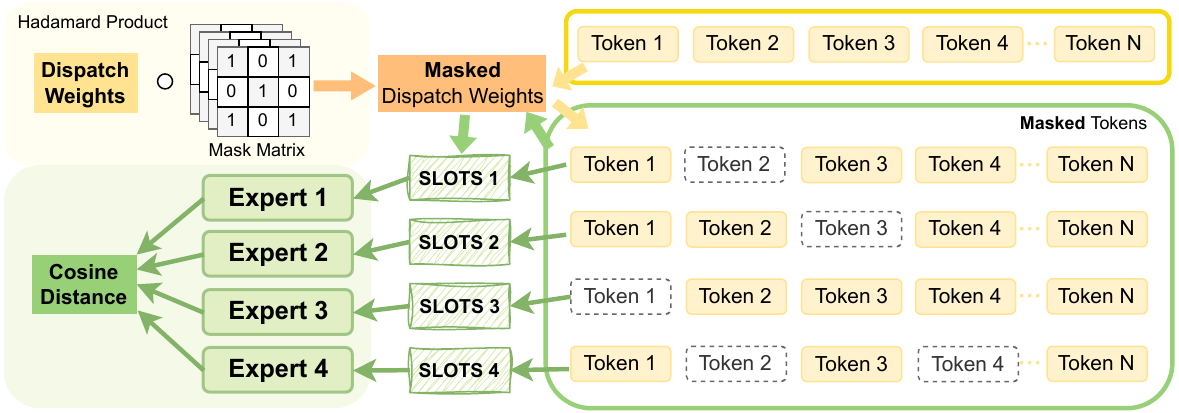} 
\caption{The Masked Mechanism and Cosine Distance Loss of MMoE. After masking the Dispatch Weights with Mask Matrix, the input tokens calculate with it to mask a part of the information, and output the slots. The slots would be input into the expert, then the cosine distance loss calculates the similarity between the output of each expert.}
\label{fig3: MMoE}
\end{figure*}

\subsection{Sub-Protocols of IAS-CDT Protocol}
Three sub-protocols are defined in the IAS-CDT Protocol to evaluate the three kinds of generalization performance of the models.
Figure~\ref{fig: samples} shows the Train / Test usage of each dataset of the IAS-CDT Protocol.

\noindent
\textbf{Protocol-1} evaluates the model's average performance in iris anti-spoofing, including the training and testing sets of all datasets.

\noindent
\textbf{Protocol-2} evaluates the model's generalization performance of cross-racials.
The training sets include H and F, and testing sets include all subsets, because the iris samples in H and F are totally collected from Asians, while iris samples in other subsets are from multiple races.

\noindent
\textbf{Protocol-3} evaluates the model's generalization performance of cross-devices, with six sub-protocols.
The datasets are divided into six parts based on the devices used for collection.
In each sub-protocol, five parts are used to train, and the remaining one is used to test.



\section{Proposed Method} \label{Section: MMoE}
We propose this Masked-MoE anti-spoofing method (shorten as MMoE in the following) to improve the capability of iris recognition system on differentiating live irises and irises with contact lenses.
The proportion of texture differences between live irises and those with contact lenses is relatively small in iris anti-spoofing tasks, which cause the difficulties on distinguishing them in practical application scenes.
The proposed MMoE contains two main components, mask mechanism and cosine distance loss, which could improve the generalization and reduce the overfitting problem.

Because of the variety of iris collection devices, the iris recognition models would perform weak on cross-device scenes.
To improve the cross-domain capability of anti-spoofing model, we introduce MoE to alleviate the negative impacts caused by cross-domain datasets.
While MoE improves model generalization, the increased parameters combined with limited discriminative features may lead to overfitting.
To address this, a masking mechanism is introduced to guide experts in focusing on critical features for anti-spoofing decisions by dropping out redundant information.
And then, to mitigate gaps between experts handling masked and unmasked features, comparisons are made between their outputs, leveraging complementarity among experts to enhance overall model generalization.

\begin{figure}[t]
\centering
\includegraphics[width=\columnwidth]{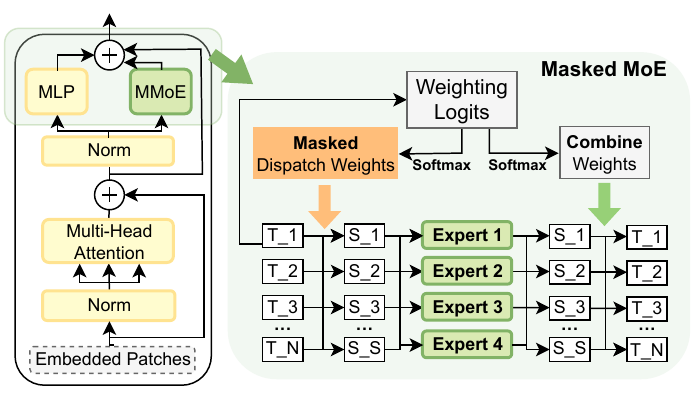} 
\caption{This figure shows the combination framework of MMoE with CLIP. The MMoE block is parallel with the MLP block in the image encoder of the CLIP.}
\label{fig: clip+moe}
\end{figure}

Figure~\ref{fig3: MMoE} demonstrates the critical part of the framework, and Algorithm~\ref{algorithm: maskedmoe} provides the working process, including the details of masking dispatch weights and the measure of cosine distances between experts.
Compared with soft MoE~\cite{puigcerver2023sparse}, our MMoE masked parts of the features before putting them into experts and calculating the cosine distances of the output of different experts. This MMoE is easily to be added to the image encoder of CLIP~\cite{clip} as shown in the Figure~\ref{fig: clip+moe}.

\subsection{Mask Mechanism of MMoE}
As mentioned before, MoE could improve generalization but risks overfitting due to increased parameters and insufficient discriminative features.
This overfitting problem of MoE methods inherits exists and is hard to solve~\cite{acena2022minimally, chang2013oversampling, allingham2021sparse, xie2023moec}.
Inspired by MAE~\cite{he2022masked}, which improves the generalization of auto-encoder architecture by masking patches of samples, we designed a masking mechanism where different experts focus on masked portions of features, guiding the model to learn key anti-spoofing characteristics.
This novel approach Mask Mechanism, which is different from the sample-wise masking manner of MAE, it is a feature-wise Mask Mechanism that masks parts of the latent features of images by tokens.
To achieve that, we generate a random Mask Matrix with fixed mask rates, then use it to mask the Dispatch Weights by Hadamard product.
Then the input tokens are weighted by Masked Dispatch Weights, and thus parts of them are masked as shown in Figure~\ref{fig3: MMoE}.
Moreover, considering the limited and subtle differences between live irises and those with contact lenses, the mask rate ought to be in a reasonable range, which could not only reduce the overfitting but also avoid omitting the obvious difference between liven irises and the fake ones.

We first introduce the Dispatch Weights shown in Figure~\ref{fig3: MMoE}. Equation~\ref{dispatch_weights} is the definition of Dispatch Weights $\mathbf{D}$ and the weighting process of MMoE's input.

\begin{equation}
    \begin{aligned}
    \mathbf{D}_{ij}=\frac{\exp((\mathbf{X}\boldsymbol{\Phi})_{ij})}{\sum_{{i^{\prime}=1}}^{n}\exp((\mathbf{X}\boldsymbol{\Phi})_{{i^{\prime}j}})},\quad&\tilde{\mathbf{X}}=\mathbf{D}^{\top}\mathbf{X}
    \label{dispatch_weights}
    \end{aligned}
\end{equation}

 This Dispatch Weights is the original one in Soft MoE.
 $\mathbf{X}\in\mathbb{R}^{{n\times d}}$ is the input tokens of MMoE, $\textit{n}$ denotes the number of tokens and $\textit{d}$ denotes their feature dimension.
 In MMoE with $\textit{e}$ experts, each expert deals with $\textit{s}$ slots where each slot has a corresponding vector of parameters with $\textit{d}$-dimension.
 $\mathbf{\Phi}\in\mathbb{R}^{{d\times(e\cdot s)}}$ is a randomly initialized vector to weight the input tokens. $\mathbf{D}\in\mathbb{R}^{{n\times(e\cdot s)}}$ is the dispatch weights, which is the result of applying a softmax on the columns of $\mathbf{X}\mathbf{\Phi}$.
 $\mathbf{D}_{ij}$ denotes the element in dispatch weights $\mathbf{D}$ located in the ${i}^{th}$ column and ${j}^{th}$ row.
 $\tilde{\mathbf{X}}\in\mathbb{R}^{(e\cdot s)\times d}$ is the input of Experts, a set of slots. It is obtained by weighting $\mathbf{X}$ with the dispatch weights $\mathbf{D}$.

Equation~\ref{masked_weights} denotes the main contribution of our Mask Mechanism. It shows the computation process of masking the dispatch weights.
The scale of our proposed Mask Matrix is the same scale as the one of Dispatch Weights $\mathbf{D}$, $\mathbf{M}\in\mathbb{R}^{{n\times(e\cdot s)}}$, which consists only of 0 and 1. The mask rate denotes the proportion of 0 in Mask Matrix.

\begin{equation}
    \begin{aligned}
    \widetilde{\mathbf{M^{\prime}}}_{\mathrm{xy}}=
    \begin{cases}
        0, & \mathrm{~if~}0\leq\mathrm{y<r\times n\times s} \\
        1, & \mathrm{~else}
    \end{cases}
    \label{mask_matrix}
    \end{aligned}
\end{equation}

Equation~\ref{mask_matrix} denotes the element values of the original Mask Matrix $\widetilde{\mathbf{M^{\prime}}}\in\mathbb{R}^{{n\times e\times s}}$.
The $r$ denotes the masking rate of the corresponding expert's input. The $x$ and $y$ denotes the location of the elements in matrix $\widetilde{\mathbf{M^{\prime}}}$, where $x \in [0, e)$ and $y \in [0, n \cdot s)$, and both of them are natural numbers.
And then, the elements in $\widetilde{\mathbf{M^{\prime}}}$ are shuffled randomly, $\tilde{\mathbf{M}} = shuffle(\widetilde{\mathbf{M^{\prime}}})$. Finally, the $\tilde{\mathbf{M}}$ is reshaped to fit the dimension of Dispatch Weights $\mathbf{D}$, $\mathbf{M} = rearrange(\tilde{\mathbf{M}},e,n,s -> n,(e \cdot s))$.
In practice, we employ 4 experts in each block of the backbone of CLIP's image encoder, i.e. $e = 4$.
Only one expert could get the complete information of all slots, while others could only get the masked slots.

\begin{equation}
    \begin{aligned}
    \mathbf{D'}_{ij}=\mathbf{D}_{ij}*\mathbf{M}_{ij},\quad&\tilde{\mathbf{X}}=\mathbf{D'}^{\top}\mathbf{X}
    \label{masked_weights}
    \end{aligned}
\end{equation}

And then our Masked Dispatch Weights $\mathbf{D'}$ is the Hadamard product of Dispatch Weights $\mathbf{D}$ and Mask Matrix $\mathbf{M}$.
$\mathbf{D'}_{ij}$ and $\mathbf{M}_{ij}$ denotes the element in masked dispatch weights $\mathbf{D'}$ and Mask Matrix $\mathbf{M}$ located in the ${i}^{th}$ column and ${j}^{th}$ row.
$\mathbf{X}$ is then weighted by $\mathbf{D'}$ and we get $\tilde{\mathbf{X}}$, the input of Experts.

\begin{algorithm}[t]
    \caption{$Masked-MoE$}
    \begin{algorithmic}[1]
        \Procedure{\textbf{MMoE}}{$X,\ e,\ s$}
        \State $n, d \gets X.shape$
        \State $\Phi \gets randn(e,\ s,\ d)$
        \State $logit \gets einsum(n\ d,\ e\ s\ d ->\ n\ e\ s,\ X,\ \Phi)$
        \State $D \gets logits.softmax(dim = 0)$
        \State $M \gets ones([e,\ n,\ s])$
        \State $\textbf{for}\ \ i\ $ in $\ range(e): $
        \State $ \qquad \textbf{if}\ \ i\ !=\ 0:$
        \State $\qquad \qquad M[i][0:n*s*0.1] = 0$
        \State $\qquad   \textbf{endif}$
        \State $\qquad$   shuffle$(M[i])$
        \State $\textbf{endfor}$
        \State $D \gets D\ *\ M$
        \State $C \gets rearrange(logits, \  n\ e\ s\ ->\ n\ (e\ s))$
        \State $C \gets C.softmax(dim = 1)$
        \State $slots \gets einsum(n\ d,\ n\ e\ s\ ->\ e\ s\ d,\ X,\ D)$
        \State $out \gets experts(slots)$
        \State $\textbf{for}\ \ i\ $ in $\ range(e): $
        \State $ \qquad \textbf{if}\ \ i\ !=\ 0:$
        \State $\qquad \qquad mmoe\_loss = cosine(out[0],out[i])$
        \State $\qquad   \textbf{endif}$
        \State $\textbf{endfor}$
        \State $out \gets rearrange(out,\ e\ s\ d\ ->\ (e\ s)\ d)$
        \State $out \gets einsum(s\ d,\ n\ s\ ->\ b\ n\ d,\ out,\ C)$
        \State \textbf{return} $out$
        \EndProcedure
    \end{algorithmic}
    \label{algorithm: maskedmoe}
\end{algorithm}

A proper mask rate can alleviate overfitting; however, since the texture differences between live irises and those with contact lenses occupy only a small proportion of the image, the masking ratio should be carefully controlled in practice to avoid obscuring these key differences and misleading the model's learning process.
Specifically, if there is no mask rate in the model, the overfitting problem would impact the anti-spoofing result negatively; if the mask rate is too large, the performs would gain negative impact as well because of the dropout of critical difference between liveness and spoofing samples.

\subsection{Cosine Distance Loss between Experts}
There are two main reasons to contribute this cosine distance loss functions. One is the inherent property of MoE, and the other one is caused by the used of Mask Mechanism.
The efficiency of MoE methods is mainly due to the characteristic sparseness, which means that each expert could learn different knowledge~\cite{cai2024survey}.
But the knowledges between each expert are not shared, which would limit the improvement of generalization brought by MoE on iris recognition models.
So, if we could find a way to let experts learn the knowledge from others, it may improve the generalization of the models.
Otherwise, the mask mechanism introduces gaps between the outputs of experts handling masked and unmasked features, so we employ this cosine distance loss to mitigate these gaps.

To address these problems, based on the Mask Mechanism and Soft MoE, we employ this cosine distance loss, which could calculate the distance between the output of experts, to encourage each expert to guess the unseen part and make up the masked information.
Specifically, the outputs of masked-feature experts are compared with the one of unmasked-feature experts, and their complementarities are leveraged to enhance the model's overall generalization.

\begin{equation}
    \begin{aligned}
        \mathbf{C}_{ij}=\frac{\exp((\mathbf{X}\boldsymbol{\Phi})_{ij})}{\sum_{{j^{\prime}=1}}^{n}\exp((\mathbf{X}\boldsymbol{\Phi})_{{ij^{\prime}}})},\quad&\mathbf{Y}=\mathbf{C}\tilde{\mathbf{Y}}
    \label{combine_weights}
    \end{aligned}
\end{equation}

In SoftMoE, a combine-weights is defined to combine the output slots of experts, and its definition is shown in Equation~\ref{combine_weights}. The combine-weights $\mathbf{C} \in \mathbb{R}^{{d\times(e\cdot s)}}$ is with the same dimention as $\mathbf{D}$, shown in Equation~\ref{dispatch_weights}.
The $\tilde{\mathbf{Y}}\in \mathbb{R}^{{(e\cdot s)}\times d}$ is the output of experts, and its computing process is shown in Equation~\ref{Experts_output}.

\begin{equation}
    \begin{aligned}
    \tilde{\mathbf{Y}}=Experts(\tilde{\mathbf{X}})
    \label{Experts_output}
    \end{aligned}
\end{equation}

Compared with the original SoftMoE, the cosine distance loss is employed before the process of Equation~\ref{Experts_output}.
We select the concept of cosine distance as the backbone of the loss function instead of Mean Squared Error (MSE).
That is due to the aim of MoE is to measure the absolute error between two tensors, which is not suitable for evaluating the similarity between the outputs of experts.
Thus, we choose the cosine distance to calculate the similarity between the output of each expert and not request the same.
In detail, we masked three experts; the remaining one could see all of the information and request three masked experts' output similar to the remaining one.
As shown in Algorithm~\ref{algorithm: maskedmoe}, all the outputs of the masked experts request to calculate the cosine distance with the unmasked one.

Equation~\ref{cosine_distance} denotes the calculation process.

\begin{equation}
    \begin{aligned}
    cosine\_distance_k=\frac{1}{s}\sum_{j=1}^{s}\sum_{i=2}^{e}\left(1-\frac{\tilde{Y^{1}_{j}}\tilde{Y^{i}_{j}}}{\|\tilde{Y^{1}_{j}}\|\|\tilde{Y^{i}_{j}}\|}\right)
    \label{cosine_distance}
    \end{aligned}
\end{equation}

$\tilde{Y^{i}_{j}}$ denotes the $j^{th}$ slot of the $i^{th}$ Expert's output. The Cosine Distance is the mean of the cosine similarity between the $j^{th}$ slots of the two Experts. The $e$ is the number of Experts and the $s$ is the number of slots. The $cosine\_distance_k$ denotes the $k^{th}$ residual block of ViT-B/16, which is the backbone of our image encoder.

As shown in Equation~\ref{mmoe_losss}, the $cosine\_distance$ is used to calculate our mmoe loss $L_{mmoe}$.

\begin{equation}
    \begin{aligned}
    L_{mmoe}=\frac{1}{12}\sum_{k=1}^{12}cosine\_distance_k
    \label{mmoe_losss}
    \end{aligned}
\end{equation}

The ViT-B/16 consists of 12 residual blocks. Our mmoe loss, $L_{mmoe}$ is mean of the 12 blocks' Cosine Distance.
This $L_{mmoe}$ is added together with $L_{clip}$, which is the loss function of CLIP as shown in Equation~\ref{CLIPLOSS}.

\begin{equation}
    \begin{aligned}
    L_{clip}=-\frac{1}{2N}\sum_{i=1}^N\left(\log\frac{\exp(S_{ii})}{\sum_{j=1}^N\exp(S_{ij})}+\log\frac{\exp(S_{ii})}{\sum_{j=1}^N\exp(S_{ji})}\right)
    \label{CLIPLOSS}
    \end{aligned}
\end{equation}

The loss function of CLIP model is a cross entropy between images and texts. The $\mathbf{S}$ is the similarity matrix and the element $S_{ij}$ of the similarity matrix $\mathbf{S}$ denotes the similarity between the $i^{th}$ image embedding and the $j^{th}$ text embedding.

\begin{equation}
    \begin{aligned}
    L_{mmoe\_with\_clip}=L_{clip} +L_{mmoe}
    \label{LOSS}
    \end{aligned}
\end{equation}

All of the loss function of our proposed MMoE with CLIP is shown as Equation~\ref{LOSS}.

\begin{figure}[t]
\centering
\includegraphics[width=0.95\linewidth]{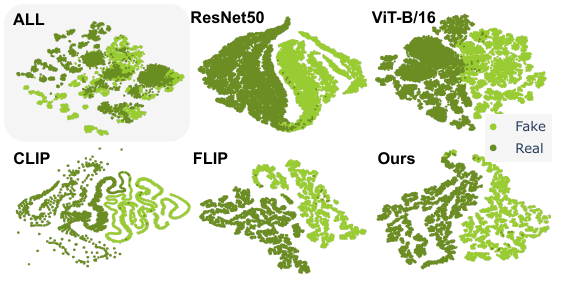} 
\caption{The feature distribution of each method on Protocol-1. Our proposed method has the clearest border and regular shape.}
\label{fig: distributions}
\end{figure}

\begin{table*}[t]
\fontsize{9pt}{10pt}\selectfont
\setlength{\tabcolsep}{1pt}
\renewcommand{\arraystretch}{1.2}
\centering
\begin{tabular}{c|cccc|cccc|cccc}
\hline

& \multicolumn{4}{c|}{\textbf{Protocol-1}}
& \multicolumn{4}{c|}{\textbf{Protocol-2}}
& \multicolumn{4}{c}{\textbf{Protocol-3 (Average)}} \\

Method
& ACER(\%)$\downarrow$  & ACC(\%)$\uparrow$ & AUC(\%)$\uparrow$ & EER(\%)$\downarrow$
& ACER(\%)$\downarrow$  & ACC(\%)$\uparrow$ & AUC(\%)$\uparrow$ & EER(\%)$\downarrow$
& ACER(\%)$\downarrow$  & ACC(\%)$\uparrow$ & AUC(\%)$\uparrow$ & EER(\%)$\downarrow$ \\
\hline
\hline
ResNet50
& 1.95   & 98.05  & 99.56  & 1.95
& 12.38  & 87.54  & 94.97  & 12.39
& 9.81  & 90.18  & 93.45  & 11.62  \\

VIT-B/16
& 2.77      & 97.23   & 99.53  & 2.77
& 12.37   & 87.63   & 95.57   & 12.35
& 12.67     & 87.34    & 91.28    & 12.31      \\

CLIP
& 1.41     & 98.59    & 99.61    & 1.41
& 8.09    & 91.91    & 97.82    & 8.11
& 6.85     & 93.14     & 95.53    & 8.97    \\

FLIP
&  1.45  &  98.55  &  \textbf{99.77}  & 1.45
&  9.25  &  90.72  &  96.95  &  9.18
& 5.83 & 94.17 & \textbf{97.58} & 5.81 \\

\hline

\textbf{Ours}
& \textbf{1.23}       & \textbf{98.76}        &   99.34    & \textbf{1.18}
&   \textbf{7.59}   & \textbf{92.41}        & \textbf{98.14}        &   \textbf{7.59}
&   \textbf{4.85}   &   \textbf{95.12}   &    97.26   &  \textbf{5.68}     \\

\hline

\end{tabular}
\vspace{1.0em}
\caption{This table shows the evaluation results on Protocol-1, Protocol-2, and Protocol-3 (Average).}
\label{Tab: experiment-main}
\end{table*}

\begin{table*}[ht]
\fontsize{9pt}{10pt}\selectfont
\setlength{\tabcolsep}{1pt}
\renewcommand{\arraystretch}{1.2}
\centering
\begin{tabular}{c|cccc|cccc|cccc}
\hline
& \multicolumn{4}{c|}{\textbf{Protocol-3.1}}
& \multicolumn{4}{c|}{\textbf{Protocol-3.2}}
& \multicolumn{4}{c}{\textbf{Protocol-3.3}} \\

Method
& ACER(\%)$\downarrow$  & ACC(\%)$\uparrow$ & AUC(\%)$\uparrow$ & EER(\%)$\downarrow$
& ACER(\%)$\downarrow$  & ACC(\%)$\uparrow$ & AUC(\%)$\uparrow$ & EER(\%)$\downarrow$
& ACER(\%)$\downarrow$  & ACC(\%)$\uparrow$ & AUC(\%)$\uparrow$ & EER(\%)$\downarrow$ \\
\hline
\hline
ResNet50
&  0.67  &   99.33   &   \textbf{99.98}   &   0.68
&   21.95   &   78.05   &   87.02  &  21.94
&   13.98   &   86.02   &  93.51  &   13.94   \\

VIT-B/16
&    1.30    &    98.74    &   99.57   &    1.35
&    27.56    &    72.44    &   78.02    &  27.56
&    17.23    &   82.77     &    90.57    &   17.23   \\

CLIP
&     1.34     &    98.66    &     99.94    &     1.35
& 19.60    &    80.39    &     88.46    &      19.64
&   \textbf{6.21}       &    \textbf{93.79}    &   97.97   &   \textbf{6.19 }     \\

FLIP
&  1.39  &  98.59  &   99.91 & 1.35
&  15.54  & 84.46 &  91.28  & 15.53
&  8.83 &  91.17  &  96.78  &  8.80  \\

\hline
\textbf{Ours}
&   \textbf{0.29}   &   \textbf{99.48}  &   99.96   & \textbf{0.00}
&  \textbf{15.18}  &   \textbf{84.83}   &  \textbf{91.97}  &  \textbf{15.13}
&  6.34  &  93.66  &  \textbf{98.01}  &  6.34 \\

\hline
& \multicolumn{4}{c|}{\textbf{Protocol-3.4}}
& \multicolumn{4}{c|}{\textbf{Protocol-3.5}}
& \multicolumn{4}{c}{\textbf{Protocol-3.6}}         \\

Method
& ACER(\%)$\downarrow$  & ACC(\%)$\uparrow$ & AUC(\%)$\uparrow$ & EER(\%)$\downarrow$
& ACER(\%)$\downarrow$  & ACC(\%)$\uparrow$ & AUC(\%)$\uparrow$ & EER(\%)$\downarrow$
& ACER(\%)$\downarrow$  & ACC(\%)$\uparrow$ & AUC(\%)$\uparrow$ & EER(\%)$\downarrow$ \\
\hline
\hline
ResNet50
& 20.87   & 79.13   & 80.26   & 31.98
& 0.97    & 98.97   & \textbf{99.93}   & 0.77
& 0.43    & 99.57   & 99.99   & 0.40  \\

VIT-B/16
& 26.71     & 73.29    & 79.94     & 24.62
& 1.03     & 98.97    & 99.92     & 0.92
& 2.19     & 97.81    & 99.68     & 2.20    \\

CLIP
& 13.19    & 86.81    & 87.07    & 25.58
& 0.49     & 99.50    & 99.72    & 0.77
& 0.29     & 99.71    & 99.79    & 0.30    \\

FLIP
& 8.30 & 91.70 &  \textbf{97.73}  & \textbf{8.28}
&  0.52  & 99.47 & 99.76  & 0.62
&  0.38  & 99.62 & 99.99 &  0.30   \\

\hline
\textbf{Ours}
&  \textbf{6.57}  &  \textbf{93.43}  &   93.90   &  11.94
&  \textbf{0.50}  &  \textbf{99.50}  &  99.69  &  \textbf{0.46}
&  \textbf{0.19}  &  \textbf{99.81}  &  \textbf{100.00}  & \textbf{0.20}  \\
\hline

\end{tabular}
\vspace{1.0em}
\caption{This table shows the specific evaluation results of each sub-protocol in Protocol-3.}
\label{Tab: experiment-protocol3}
\end{table*}



\section{Experiment}
\label{Section: Experiment}

\subsection{Datasets and Experimental Setting}
In this section, we re-state the dataset and the application of three sub-protocols of the Cross-Testing Protocol, and introduce some experimental settings, containing the evaluation metrics, the device, and other details of experiments.
The datasets we employed in the experiments are those involved in our proposed Cross-Testing Protocol. For more details of the Cross-Testing Protocol, Table~\ref{Tab: datasets} lists more information.
Each of three sub-protocols of the Cross-Testing Protocol has been evaluated on, to respectively confirm three different performances of iris anti-spoofing models, including our proposed MMoE. Protocol-1 evaluates the average performance of models, and Protocol-2 and Protocol-3 evaluate the cross-racial adaption and cross-device adaption respectively.
We choose the Average Classification Error Rate (ACER), the Overall Detection Accuracy (ACC), the Area Under the Curve (AUC), and the Equivalent Error Rate (EER) as performance evaluation metrics.
The experimental details are as follows:
The backbone of CLIP's image encoder is ViT-B/16.
The mask rate is set as "0, 0.1, 0.1, 0.1.", and the prompts are set as \textit{"This is an image of \textless label\textgreater eye."}
We employ NVIDIA A100 GPUs to train and test with the Adam
 optimizer and the original learning rate is $10^{-6}$.

\subsection{Performance Evaluation}
 To verify the effectiveness of our proposed MMoE, some powerful deep learning models are also selected to be tested. The ResNet50~\cite{he2016deep}, ViT-B/16~\cite{ViT}, CLIP~\cite{clip}, and FLIP~\cite{srivatsan2023flip} are chosen as representative models, then benchmarked them with  ACER, ACC, AUC, and EER. And their classification borders in latent space are visualized as well.

\begin{figure}[t]
\centering
\includegraphics[width=0.95\columnwidth]{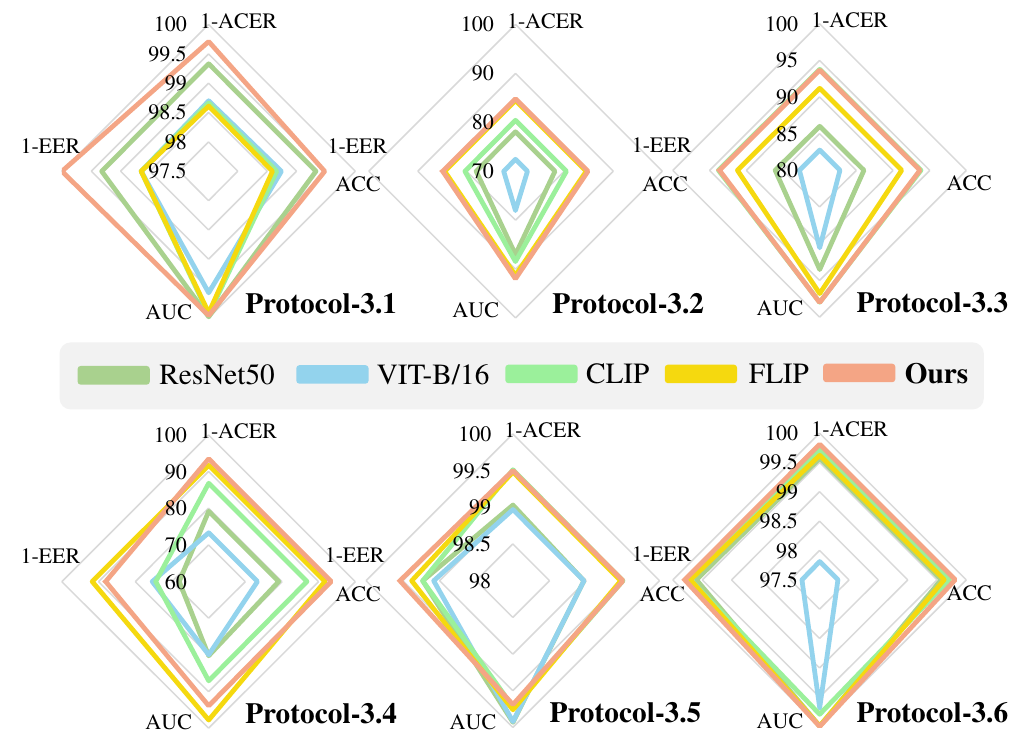}
\caption{This figure shows the ACER, ACC, AUC, and EER performed by ResNet50, ViT-B/16, CLIP, FLIP, and Ours conducted on the 6 sub-protocols of the Protocol-3.}
\label{fig: protocol-3}
\end{figure}

The experimental result of the basic performances of all models are shown in Table~\ref{Tab: experiment-main} and Table~\ref{Tab: experiment-protocol3}.
Table~\ref{Tab: experiment-main} shows all the evaluation results including our proposed MMoE with CLIP and other models on Protocol-1, Protocol-2, and Protocol-3 (Average).
Table~\ref{Tab: experiment-protocol3} provides the specific results of each sub-protocol in Protocol-3.
To compare the performance of Ours with other methods, Figure\ref{fig: protocol-3} and Figure\ref{fig: protocol-1-2} visualize their ACER, ACC, AUC, and EER. Specifically, the Figure\ref{fig: protocol-3} shows their performance comparison on the 6 sub-protocols of Protocol-3, and the Figure\ref{fig: protocol-1-2} shows their performance comparison on Protocol-1 and Protocol-2.
From the comparison, our proposed MMoE could achieve the SOTA performance on nearly all evaluation metrics for the three protocols.

The evaluation results show that our proposed MMoE method (which is shortened as \textit{Ours} in the following) with CLIP gets the best performance on all three sub-protocols of the Cross-Testing Protocol, among those models.
For Protocol-1, Ours achieves 1.23\% ACER, which is a nearly 13\% improvement compared to the vanilla CLIP with 1.41\% ACER. It means that our MMoE method could effectively detect spoofing iris samples on all datasets in the Cross-Testing Protocol and achieve SOTA performance.
For Protocol-2, Ours achieves 7.59\% ACER, much better than other previous models.
It proves our proposed method has a high generalization ability on cross-racial challenges. Verified the effectiveness of our contributions.
For Protocol-3, Ours achieves 4.85\% ACER, which has significant advantages over other previous models.
It proves our proposed method also has a high generalization ability on cross-device challenges.

Furthermore, Figure~\ref{fig: distributions} demonstrates the feature distribution of each method on Protocol-1. Our proposed method has the clearest border and regular shape in the latent space, demonstrating its good ability in this iris anti-spoofing field.

In summary, upon all the above evaluations, our proposed MMoE method could effectively detect the spoofing iris images and perform high generalization on both cross-racial and cross-device situations. It achieves the SOTA performance on all protocols of the Cross-Testing Protocol.

\begin{table}[t]
\fontsize{10pt}{10pt}\selectfont
\setlength{\tabcolsep}{3pt}
\renewcommand{\arraystretch}{1.2}
\centering
\begin{tabular}{l|cccc}
\hline
         & \multicolumn{4}{c}{\textbf{Mask-Rate Analysis Protocol-1}}      \\
Rates  & ACER(\%)$\downarrow$ & ACC(\%)$\uparrow$ & AUC(\%)$\uparrow$ & EER(\%)$\downarrow$  \\
\hline
\hline
\textbf{Rate\_1}    & \textbf{1.23} &   \textbf{98.76} & 99.34 & \textbf{1.18} \\
Rate\_2   & 1.38 & 98.63 & \textbf{99.51} & 1.41 \\
Rate\_3 &  1.60  &  97.45  &  99.47  &  1.59  \\
Rate\_4   & 1.43   & 98.57 & 99.51 &  1.43    \\

\hline
         & \multicolumn{4}{c}{\textbf{Mask-Rate Analysis Protocol-2}}      \\
Rates  & ACER(\%)$\downarrow$ & ACC(\%)$\uparrow$ & AUC(\%)$\uparrow$ & EER(\%)$\downarrow$  \\
\hline
\hline
\textbf{Rate\_1}    &  \textbf{7.59}  & \textbf{92.14} & \textbf{98.14} &  \textbf{7.59}  \\
Rate\_2   & 8.14 & 91.86 & 97.67 &  8.13  \\
Rate\_3   & 7.95 & 92.06 & 97.57 &  7.95  \\
Rate\_4   & 8.70 & 91.30 & 97.36 &  8.72  \\

\hline

\end{tabular}
\vspace{1.0em}
\caption{The evaluation results of our MMoE with CLIP under different sets of mask-rates. The Rate\_1 has the best performance on both Protocol-1 and Protocol-2.}
\label{Tab: rate}
\end{table}

\begin{figure}[!th]
\centering
\includegraphics[width=0.95\columnwidth]{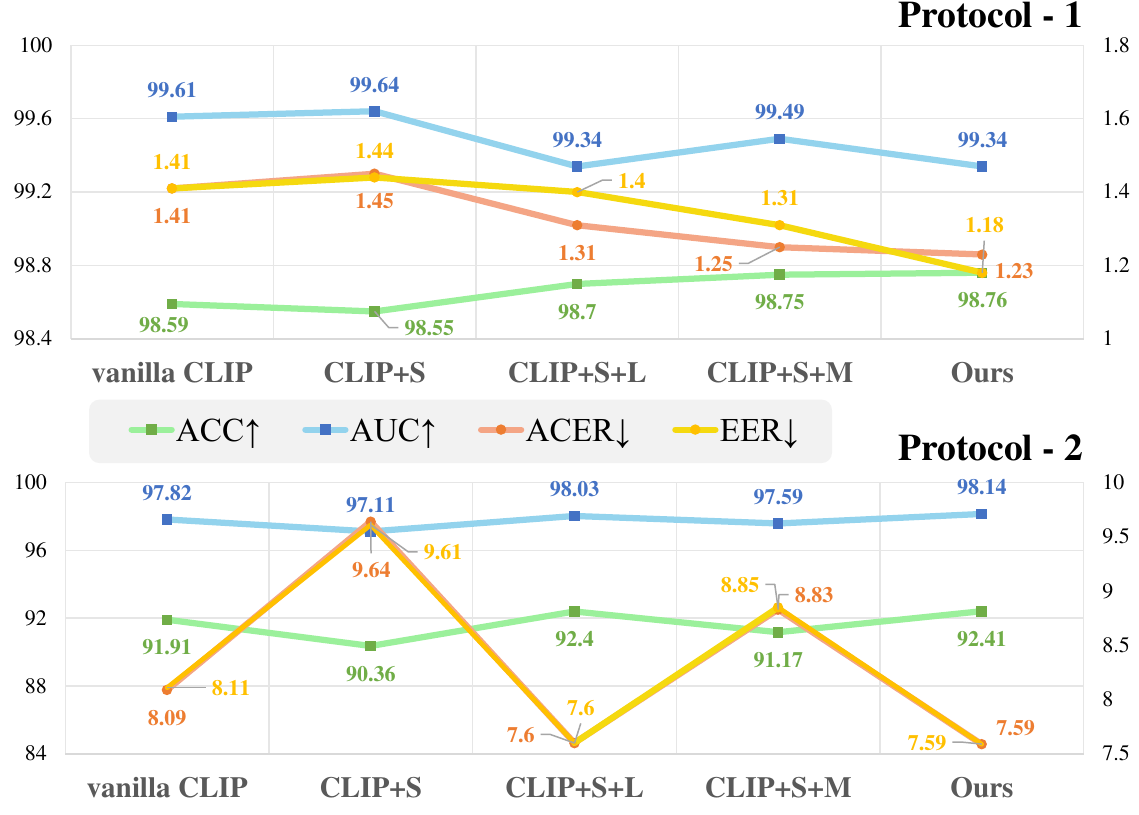}
\caption{This figure demonstrates the trends of ACERs, ACCs, AUCs, and EERs in ablation experiments, where the above one shows the results of on Protocol-1, and the below one shows the results on Protocol-2.}
\label{fig: ablation}
\end{figure}

\subsection{Mask Rate Analysis}
To discuss the impact of different rates of the mask, we set up 4 sets of mask rates and tested them on both Protocol-1 and Protocol-2.
Among those, the first two sets mask the same rate for both three experts, and one expert could see all tokens with complete information. The difference between these two sets is that the mask rate of the first set is less than the one of the second set.
The second set is to confirm that the mask rate ought to be in a reasonable range, not too large.
The third one masks different rates for three experts, while one expert could also see all information as similar to the first two sets. The third set is to evaluate how the rates of different experts impact the performance of the whole model.
The last one is quite different, in that all experts are masked parts of tokens, which means the MMoE model could not gain the complete input. This set demonstrates the significance of the unmasked feature.
The exact numbers of the mask rates in each set are shown as follows:

\textbf{Rate\_1}: 0, 0.1, 0.1, 0.1.
\textbf{Rate\_2}: 0, 0.5, 0.5, 0.5.

\textbf{Rate\_3}: 0, 0.15, 0.3, 0.45.
\textbf{Rate\_4}: 0.1, 0.1, 0.1, 0.1.

\noindent

\begin{figure}[!th]
\centering
\includegraphics[width=0.85\columnwidth]{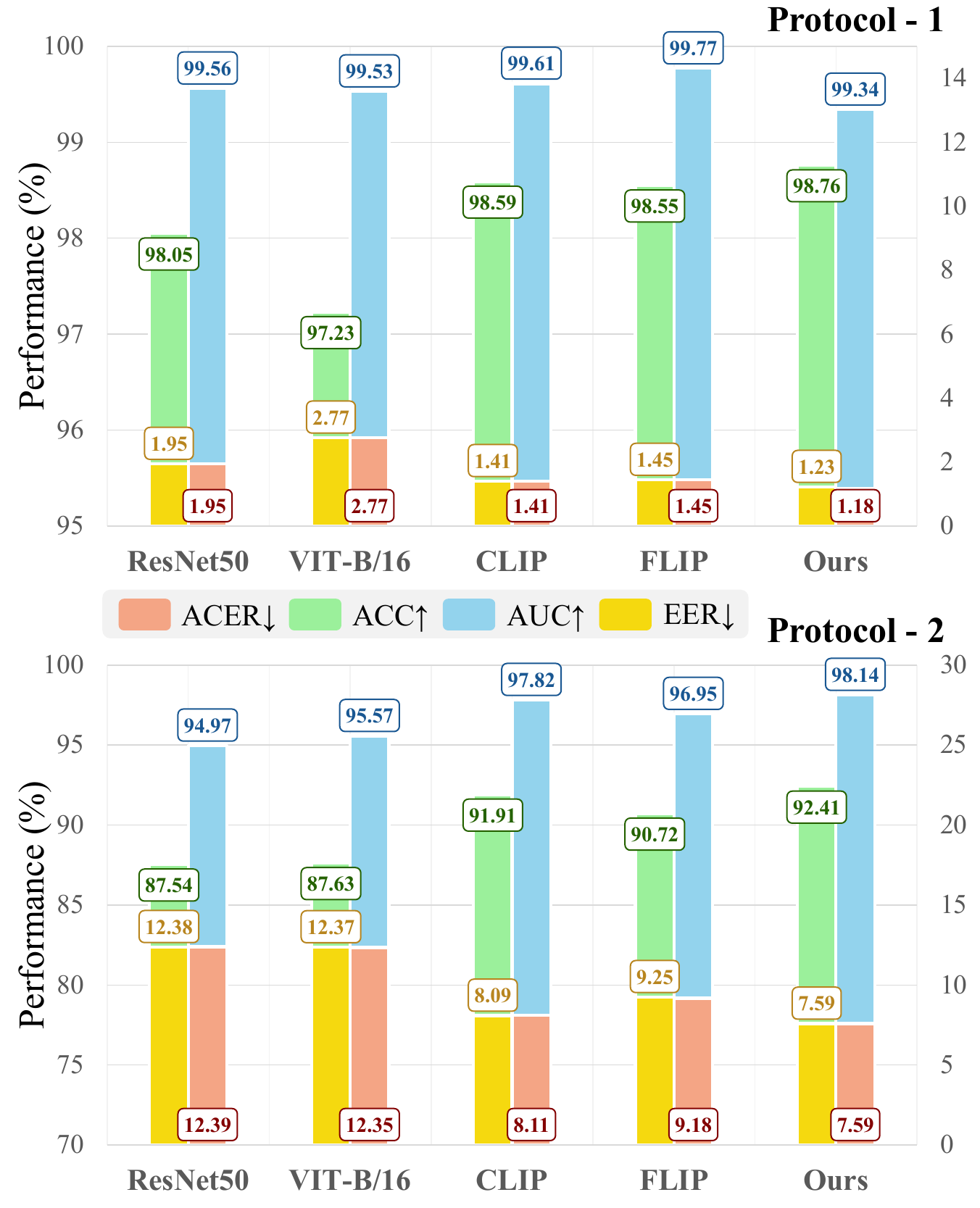}
\caption{This figure shows the ACER, ACC, AUC, and EER performed by ResNet50, ViT-B/16, CLIP, FLIP, and Ours conducted on Protocol-1 and Protocol-2,
where the above one shows the results on Protocol-2 and the below one shows those on Protocol-2.}
\label{fig: protocol-1-2}
\end{figure}

According to the evaluation results shown in Table~\ref{Tab: rate}, the result of the first set achieves the SOTA performance. And the results of other sets are weak in different degrees.
The Rate\_1 has the best performance on both Protocol-1 and Protocol-2, in which the ACERs are 1.23\% and 7.59\% respectively.
The Rate\_2 is a bit worse than the Rate\_1 may caused by the high mask rate would make experts lose too much information and then influence the learning space of the MMoE model.
The Rate\_3 with different mask rates set has the worst performance on Protocol-1. This difference may increase the gap between the learning ability of experts and then cause a negative impact on the training process.
The Rate\_4 has a middle performance among these mask rate sets on Protocol-1, a bit worse than the Rate\_1. And is the worst one on Protocol-2.
This set makes the MMoE model could not gain the complete input information, which would passively influence the learning space.

Consequently, these experimental results further confirms that the mask rate should be in a reasonable range and a complementary feature without masking is necessary.

\subsection{Ablation Study}
In ablation study part, we evaluate the effectiveness of the components of our MMoE model.
Two main contributions are proposed in our MMoE method: Mask Mechanism, and Cosine Distance Loss.
To practically confirm the effectiveness of our contributions, we provide the ablation study on Protocol-1 and Protocol-2.
In this part, the \textbf{CLIP} means the vanilla CLIP framework, \textbf{S} is SoftMoE, \textbf{L} is the cosine distance between experts, and \textbf{M} is the Mask Mechanism.
For example, \textbf{CLIP+S+M+L} means CLIP framework with SoftMoE, Mask Mechanism, and cosine distance between experts, which is briefly called MMoE with CLIP or Ours.
The evaluation results are shown in Table~\ref{Tab: ablation}, and the trends of ACERs, ACCs, AUCs and EERs are demonstrated in Figure~\ref{fig: ablation}.
It could be observed that after adding the Soft MoE "S" into the vanilla CLIP, the performance significantly declined, especially on Protocol-2, which might be caused by the overfitting problem of introducing MoE.
After adding our contributions L and M, the performances improved on both Protocols.
Specifically, to compare the ablation results, the CLIP+S+L is better on Protocol-2 which focuses on generalization, while the CLIP+S+M is better on Protocol-1 which focuses on average performance.
That proves our contributions Cosine Distance Loss "L" could improve the model's generalization ability and Mask Mechanism "M" significantly alleviates the overfitting problem caused by MoE.
The CLIP+S+M+L (Ours) shows the best performance among these combined models, the ACER is 1.23\% on Protocol-1 and 7.59\% on Protocol-2. That good performance is thanks to both alleviating overfitting and improving generalization with our two contributions.

Consequently, the experimental results show that the Maks Mechanism could truly reduce the overfitting caused by the MoE, the Cosine Distance Loss could truly improve the generalization of the whole model, and both of these two components could improve the average performance of the detection of spoofing iris.




\begin{table}[t]
\fontsize{10pt}{10pt}\selectfont
\setlength{\tabcolsep}{1.5pt}
\renewcommand{\arraystretch}{1.2}
\centering
\begin{tabular}{l|cccc}
\hline
         & \multicolumn{4}{c}{\textbf{Ablation Study Protocol-1}}      \\
Methods  & ACER(\%)$\downarrow$  & ACC(\%)$\uparrow$ & AUC(\%)$\uparrow$ & EER(\%)$\downarrow$   \\
\hline
\hline
vanilla CLIP          & 1.41-             & 98.59-             & 99.61-                     & 1.41-      \\
CLIP+S                &   1.45$\uparrow$         &  98.55$\downarrow$                  & \textbf{99.64}$\uparrow$         & 1.44$\uparrow$\\
CLIP+S+L              &  1.31$\downarrow$          & 98.70$\uparrow$                &  99.34$\downarrow$                  & 1.40$\downarrow$  \\
CLIP+S+M               &  1.25$\downarrow$          &  98.75$\uparrow$                &  99.49$\uparrow$                &  1.31$\downarrow$   \\
\textbf{CLIP+S+M+L}         & \textbf{1.23}$\downarrow$        & \textbf{98.76}$\uparrow$      &   99.34$\downarrow$           & \textbf{1.18}$\downarrow$ \\
\hline
          & \multicolumn{4}{c}{\textbf{Ablation Study Protocol-2}}  \\
Methods  & ACER(\%)$\downarrow$  & ACC(\%)$\uparrow$ & AUC(\%)$\uparrow$ & EER(\%)$\downarrow$   \\
\hline
\hline
vanilla CLIP           & 8.09-                & 91.91-                & 97.82-                             & 8.11-     \\
CLIP+S                 &  9.64$\uparrow$        &  90.36$\downarrow$   &  97.11$\downarrow$                     &  9.61$\uparrow$  \\
CLIP+S+L                 &  7.60$\downarrow$    &  92.40$\uparrow$        &  98.03$\uparrow$                &  7.60$\downarrow$   \\
CLIP+S+M                & 8.83$\uparrow$        &  91.17$\downarrow$        &  97.59$\downarrow$                 &  8.85$\uparrow$  \\
\textbf{CLIP+S+M+L}   &\textbf{7.59}$\downarrow$  & \textbf{92.41}$\uparrow$     & \textbf{98.14}$\uparrow$           & \textbf{7.59}$\downarrow$        \\
\hline
\end{tabular}
\vspace{1.0em}
\caption{The evaluation results of each ablation on Protocol-1 and Protocol-2.}
\label{Tab: ablation}
\end{table}

\section{Conclusion and Future Work}
In this paper, we proposed a unified framework for iris anti-spoofing, which includes an Iris Anti-Spoofing Cross-Domain-Testing (IAS-CDT) Protocol and a Masked-MoE (MMoE) method.
The IAS-CDT Protocol provides an evaluation standard for testing the generalization of iris recognition models.
As both the theoretical analysis and the experimental confirmations demonstrates, our MMoE method could significantly improves the generalization of iris anti-spoofing models and alleviates overfitting problems.
In future work, we will focus on studying the boundary of the MMoE's adaptive capability.

\bibliographystyle{IEEEtran}
\bibliography{IEEEabrv, egbib}

\end{document}